\documentclass[letterpaper, 10 pt, conference]{ieeeconf}  

\IEEEoverridecommandlockouts                              

\usepackage{multirow}
\usepackage{amsmath}  
\usepackage{amssymb}  
\usepackage[colorlinks,urlcolor=blue,linkcolor=blue,citecolor=blue]{hyperref}
\usepackage{color,array}
\usepackage{xcolor}
\usepackage{float}
\usepackage{booktabs}
\usepackage{diagbox}
\usepackage{graphicx}
\usepackage{setspace}  
\usepackage{subcaption}
\usepackage{caption}
\DeclareCaptionLabelSeparator{periodspace}{.\quad}
\usepackage{graphics} 
\usepackage{epsfig} 
\usepackage{times} 
\usepackage{amsmath} 
\usepackage{amssymb}  
\usepackage{multirow} 
\usepackage{booktabs} 
\usepackage{gensymb}  
\usepackage{bm}  
\usepackage{underscore}  
\usepackage[linesnumbered,ruled,vlined]{algorithm2e}
\usepackage{threeparttable}

\usepackage{bbding} 
\usepackage{cite}
\usepackage{url}
\usepackage{hyperref}
\hypersetup{
    colorlinks  = true,
    citecolor = blue,
    linkcolor = blue,   
    urlcolor = blue
}

\title{\LARGE \bf
Monocular Person Localization under Camera Ego-Motion
}

\author{Yu Zhan$^{1}$, Hanjing Ye$^{1}$ and Hong Zhang$^{1*}$
\thanks{*corresponding author (hzhang@sustech.edu.cn).}
\thanks{$^{1}$Yu Zhan, Hanjing Ye and Hong Zhang are with Shenzhen Key Laboratory of Robotics and Computer Vision, Southern University of Science and Technology (SUSTech), and the Department of Electronic and Electrical Engineering, SUSTech.}
\thanks{
This work was supported in part by Shenzhen Science and Technology Program (No. SGDX20240115111759002), by Meituan and by SUSTech, China (G03034K003).
}
}

\begin{document}

\maketitle
\thispagestyle{empty}
\pagestyle{empty}

\begin{abstract}
Localizing a person from a moving monocular camera is critical for Human-Robot Interaction (HRI). To estimate the 3D human position from a 2D image, existing methods either depend on the geometric assumption of a fixed camera or use a position regression model trained on datasets containing little camera ego-motion. These methods are vulnerable to severe camera ego-motion, resulting in inaccurate person localization. We consider person localization as a part of a pose estimation problem. By representing a human with a four-point model, our method jointly estimates the 2D camera attitude and the person's 3D location through optimization. Evaluations on both public datasets and real robot experiments demonstrate our method outperforms baselines in person localization accuracy. Our method is further implemented into a person-following system and deployed on an agile quadruped robot. 

\end{abstract}

\section{INTRODUCTION}
\begin{figure}[t]
        \centering
        \includegraphics[width=\linewidth, height=0.55\linewidth]{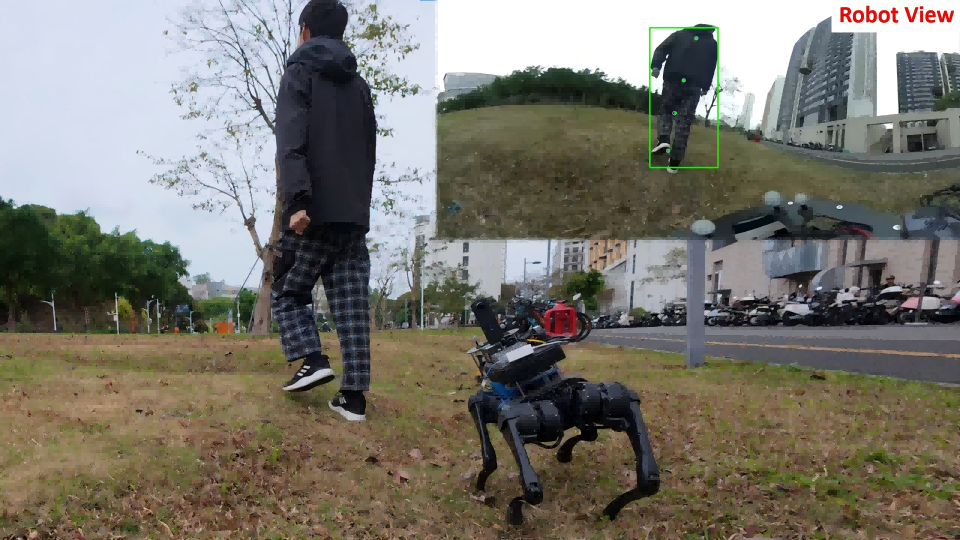}
        \caption{A scenario of a quadruped robot following a person through a rugged lawn. The robot view is from an onboard panoramic camera (see Sec. \ref{platform-text}). The robot's dynamic motion induces severe camera ego-motion and vibration, which bring challenges for person localization.}
        \label{introduction}
        \vspace*{-0.30in}
\end{figure}

Person localization is critical for robotic applications. In practice, accurate and stable person localization not only lays the foundation for robot planning and control in human-robot interaction tasks such as Robot Person Following (RPF) \cite{islam2019person}, but is also important for egocentric crowd behavior analysis and anomaly detection\cite{anomaly}. In recent years, many methods have been developed for monocular person localization from a moving camera\cite{koide2020monocular, ye2023robot, Bacchin-Italian}, most of which rely on the restrictive assumption that the camera is mounted at a fixed height and attitude (roll and pitch angles) from the ground plane\cite{islam2019person}. However, many real-world robotic deployments starkly break this assumption, such as autonomous golf caddies moving on undulating golf courses\cite{caddy} and quadruped robots traversing rough terrains\cite{unitree}. Fig.~\ref{introduction} presents our target scenario: a quadruped robot traversing a rugged lawn while following a person. In such challenging environments, the robot's severe ego-motion poses challenges to stable person localization and tracking.

Practically, to localize a person in 3D while avoiding ego-motion perturbations, many robotic applications resort to distance sensors such as UWB\cite{unitree}, LiDAR\cite{zhang2021efficient, perception_engine, Multi-modal, 3D-pose-korean, leg_tracker}, and RGB-D camera\cite{zhang2021efficient}. However, using a monocular camera to realize person localization is worth researching because of its low physical complexity and cost.  

Another widely adopted solution against ego-motion is to estimate the robot state\cite{roychoudhury2023perception} to compensate for the camera ego-motion based on IMU\cite{3D-pose-korean, IJSR}, leg kinematic sensors\cite{perception_engine}, or odometry\cite{zhang2021efficient, aso2021portable} to get the current camera attitude and transform the estimated human location to a stabilized robot frame. However, robust and low-drift state estimation is challenging for a highly dynamic robot such as a quadruped, especially for IMU-based odometry. Specifically, the repetitive foot-ground impacts during a trotting gait induce significant odometry drift\cite{allione2023effects_report,yang2023cerberus}, which induces accumulating errors in the person's estimated location.

Therefore, instead of relying on a possibly error-accumulated odometry, one solution is to estimate 3D person location from single-frame observation. This problem is widely researched in the computer vision community\cite{zheng2023deep,mono-hpe-survey2} and the autonomous driving community\cite{its-survey}, including monocular depth estimation\cite{depth_anything,unidepth}, human localization\cite{monoloco++}, 3D human joint estimation\cite{moon} and human body mesh recovery\cite{multi-hmr2024}. However, most of these methods are based on deep learning; they perform based on the distribution of training data and lack generalizability to scenarios, image projection, or camera ego-motion. Another type of method represents a human with non-rigid models like SMPL\cite{smplify} and optimizes the 6-DoF pose of a person's root joint along with posture and shape parameters. However, their complex non-linear systems with numerous parameters bring challenges to highly real-time applications. We propose to estimate 3D person location from every single image shot. Once per-frame person localization achieves both high precision and real-time efficiency, it enables robust human tracking even under severe camera ego-motion.

In this paper, we propose an optimization-based method for person localization from single-frame observation. In each frame, we represent the human body as a four-point model. Based on the 2D-3D correspondence of the four points, we simultaneously optimize the 2D camera attitude and the person's 3D location. Our method achieves the lowest localization error in both public datasets and our custom dataset compared to the baselines.

Furthermore, we integrate our method into an RPF system and conduct experiments on a Unitree Go1 quadruped\cite{unitree} that follows a person through a variety of rough terrains. Our method achieves accurate and stable person localization at a long-term RPF trial with severe camera ego-motion. Our code and dataset are available at \url{https://medlartea.github.io/rpf-quadruped/}.

\section{RELATED WORK}
Estimating a person's 3D location from a monocular image is an ill-posed problem due to the inherent ambiguity of 3D-to-2D projection. To address this, existing methods either localize a person using geometric constraints derived from scene priors, such as a planar ground assumption or a known pedestrian scale (Sec. \ref{geometric}), or directly estimate the person's full 6-DoF pose via optimization (Sec. \ref{optimize}) or deep learning-based regression (Sec. \ref{regress}). In this work, we focus exclusively on approaches capable of real-time deployment on robotic systems.
\subsection{\textbf{Person Localization with Geometric Model}\label{geometric}}

Existing methods model the human body as a single line segment, combined with the assumptions of a ground plane and the upright posture of humans. Based on the correspondence between the detected human joints in the image and the geometric model, these methods solve for the 3D positions. 

Fei \emph{et al.} \cite{fei} assumes that all observed people have the same known height and stand upright on the same plane. When at least two individuals are observed, they simultaneously estimate the camera's pose relative to the horizontal plane and the physical distance between people. Aghaei \emph{et al.} \cite{proxemics} assumes that the camera is fixed at a known height and a tilt angle within a specific range from the plane. By leveraging the length of the upright human torso, the scale of the model is recovered, enabling the estimation of social distance between people. The aforementioned methods are developed for surveillance perspectives, where the camera is far away above the plane. Choi \emph{et al.} \cite{choi2010multiple} proposed a framework for tracking people from a moving monocular camera. He assumes the camera is fixed, so the image has a fixed ground horizon line, making it prone to severe pitch and roll changes. Koide \emph{et al.} \cite{koide2020monocular} proposed the first framework for locating, tracking, and following a person using a monocular camera from a robot's perspective. Ye \emph{et al.} \cite{ye2023robot} extended Koide's method to adapt to partial occlusions by continuously updating an unscented Kalman filter using observable 2D joint information. Bacchin \emph{et al.} \cite{Bacchin-Italian} used an omnidirectional camera to extend Koide's method\cite{koide2020monocular} to equirectangular projection. However, although their methods achieve robust person localization on real-time robot applications, they assume the camera pose relative to the ground plane is fixed, which fails to generalize to platforms with ego-motion and restricts the methods to wheeled robots moving on flat ground.

In summary, existing monocular person localization methods based on geometric models are constrained by the assumption of a fixed camera observing on a flat ground. When the camera is mounted on an agile off-road robot, such as a quadruped, severe ego-motion can violate these geometric assumptions, leading to inaccurate person localization results.

\subsection{\textbf{Pose Optimization from Semantic Keypoints}\label{optimize}}

Recent work has explored the use of semantic keypoints to recover 6-DoF poses. These methods bridge the gap between geometric rigidity and deformable object modeling, offering promise for human pose estimation.

A seminal approach by Pavlakos \emph{et al.} \cite{6-dof} estimates 6-DoF poses of objects by combining semantic keypoints predicted by CNNs with a deformable shape model. They optimize the 6-DoF pose over the Stiefel manifold using convex relaxation and manifold optimization. While this method achieves robustness across object categories, its reliance on the general rigidity of an object model makes it difficult for non-rigid targets like humans. Pavliv \emph{et al.} \cite{drone-swarm} applies methods in \cite{6-dof} to track drone swarms using a vision-based headset. 

BodySLAM++\cite{henning2023bodyslam++} represents a significant leap by tightly coupling human pose estimation and stereo Visual Inertial Odometry (VIO) in a factor graph. It jointly estimates SMPL body parameters, camera trajectory, and inertial states via multi-sensor fusion. Through minimizing error terms dominated by human joint reprojection and also containing shape, posture, and motion priors, it optimizes all the parameters by nonlinear least squares in real time. However, while achieving metric-scale accuracy in real time, BodySLAM++ is a multi-image work that depends on extra VIO factors, making deployment costly for resource-constrained robots. Their method's performance under severe camera ego-motion is not tested.

\subsection{\textbf{Person Localization by Learning-based Regression}\label{regress}}

Recent advances in deep learning have shifted the paradigm of monocular 3D human localization from geometric constraints to data-driven regression. These methods can be broadly categorized into two groups: (1) direct regression of 3D positions from 2D keypoints and (2) end-to-end learning-based human pose estimation frameworks.

\textbf{Regressing 3D Location from 2D Joints}. A prominent line of work leverages 2D human keypoints\cite{monoloco++} or semantic patches\cite{monois3d} as intermediate representations to infer 3D positions. MonoLoco\cite{monoloco} and its extension MonoLoco++\cite{monoloco++} regress 2D skeletal keypoints to predict probabilistic 3D human locations with uncertainty, addressing the inherent depth ambiguity caused by human height variations. However, these methods heavily rely on benchmark datasets like KITTI\cite{kitti}, which captures street-level perspectives and lacks views from a robot with large-angle ego-motion, leading to poor generalization in unseen environments. As noted in a recent survey\cite{its-survey}, in most monocular 3D detection tasks for autonomous driving scenarios, only the heading angle around the up-axis (yaw angle) is considered, which causes a misalignment problem when the camera suffers ego-motion with non-zero roll/pitch angles.

\begin{figure*}
        \centering
        \includegraphics[width=0.9\linewidth]{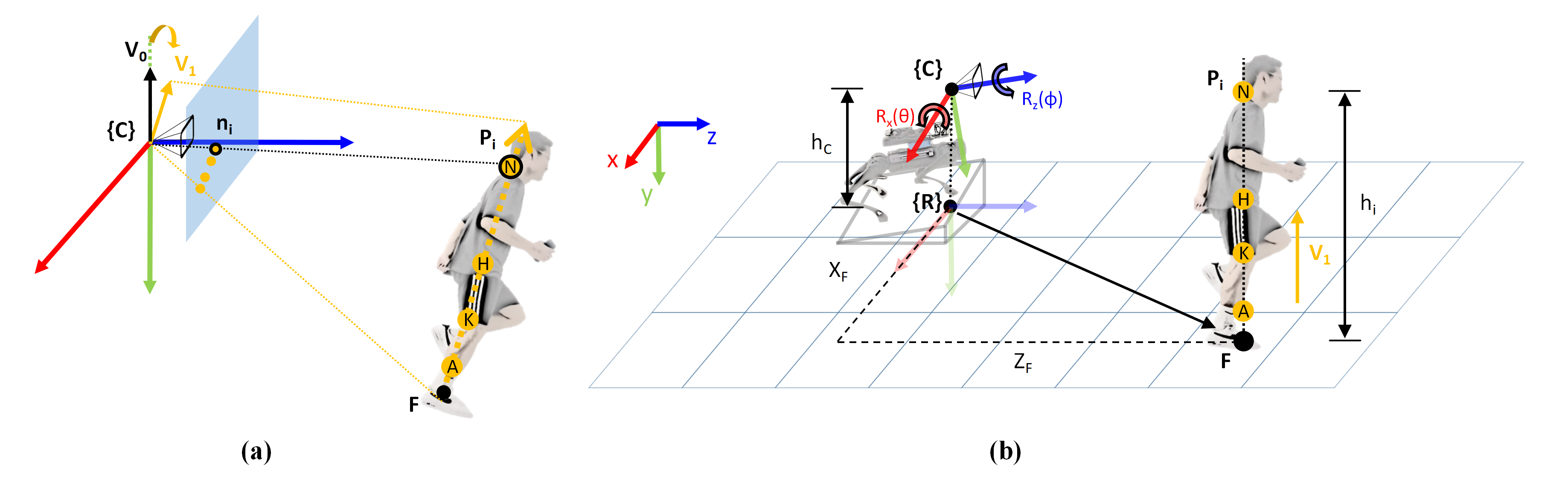}
        \caption{The geometry of our observation model. (a) In the raw camera-centric view, the person appears tilted due to the robot's ego-motion. (b) Our model assumes an upright person, representing the ego-motion as a corresponding tilt of the camera.}
        \label{observation_model}
        \vspace*{-0.15in}
\end{figure*}

\textbf{Learning-based Human Pose Estimation}. These methods aim to reconstruct a human's posture, shape, and root joint location from a single image\cite{zheng2023deep, mono-hpe-survey2}, but they often prioritize root-relative accuracy over the absolute pose relative to the camera. These approaches suffer from depth ambiguities due to the underconstrained nature of monocular inference, and their performance degrades in 'in-the-wild' settings lacking precise 3D ground-truth \cite{zheng2023deep,mono-hpe-survey2}. Moreover, their high computational cost and dependency on a pre-defined camera model with fixed intrinsics make them impractical for mobile robotic platforms. While recent advances like Multi-HMR \cite{multi-hmr2024} enable multi-person mesh and depth recovery, they are still restricted to ideal pinhole camera models. In summary, few works simultaneously address the challenges of absolute localization accuracy, real-time inference, and cross-domain generalization.

\section{METHODOLOGY}

We propose an optimization-based person localization method robust to camera ego-motion. By representing a human with a four-point model consisting of neck, hip, knee and ankle, we fit the known 3D human model into the observed 2D image measurements and simultaneously estimate the camera's 2D attitude (roll and pitch angles) along with the person's 3D location, which is inspired by the solution for PnP problems based on Bundle Adjustment. We then integrate the proposed localization method into an RPF system, achieving stable person-following behavior on a quadruped robot in rough terrain. We first introduce our human model and 2D observations in Sec. \ref{mod and obs}, then we describe our optimization details in Sec. \ref{para and cons} and Sec. \ref{optimization}. Finally, we introduce our RPF system assisted by the proposed localization method in it in Sec. \ref{framework}.

\subsection{Human Model and Observations} \label{mod and obs}

We model an upright human as a rigid body comprising four collinear points: the neck, the center of the hips, the center of the knees and the center of the ankles, which could be marked as $\mathcal{P}_{all} = \{P_{neck},P_{hip},P_{knee},P_{ankle}\}$. After projection, each point $P$ leaves the corresponding 2D point $p$ on the image plane as described in Fig. \ref{observation_model}a. The advantages of our model lie in the following three aspects: First, the projected lengths of the segments on the image plane provide a simple yet effective reflection of the person's distance from the camera; Second, the relative length ratios among these segments can indicate changes in the camera’s viewing angle, for example, when the camera is tilted downward, the upper body of a human appears stretched; Third, although the human body is deformable, $\mathcal{P}_{all}$ along the human's central axis remains relatively stable as long as the human is upright.

To detect the corresponding 2D projections of $\mathcal{P}_{all}$ on the image plane, firstly, we use YOLOX\cite{yolox} to detect people's bounding boxes and AlphaPose\cite{fang2022alphapose} to detect 2D human joints within the bounding boxes. We notice AlphaPose has good performance even under occlusion and under image distortion, as evidenced in \cite{ye2023robot}. The stable 2D joint detection lays the foundation for localization. Secondly, we derive the 2D image coordinates of the four points in $\mathcal{P}_{all}$ by computing the median of the corresponding left and right joints' image coordinates.  

After detection, we back-project the detected 2D points onto the normalized image plane. Therefore each point $P_i \in \mathcal{P}_{all}$ has its corresponding normalized image point $n_i \in \mathcal{N}_{all}$, as shown in Fig. \ref{observation_model}a. Normalization allows our method to be independent of a specific camera, unifying cameras with different intrinsics and image projections, which generalizes to different kinds of robot platforms with various camera settings.

\subsection{Parameterization and Constraints} \label{para and cons}
Since we model the human body as four collinear points, the pose of the human relative to the camera comprises 2-DoF rotation and 3-DoF translation. Specifically, we ignore the orientation of the human body around its central axis, which is not critical for localization and can be recovered independently by methods such as Part-HOE \cite{part-hoe}. This simplification makes the problem more tractable while retaining sufficient information for accurate localization. 

We assume that a human's footprint $\mathbf{F}\in\mathbb{R}^{3}$ in camera frame $\{\mathbf{C}\}$ represents his position, and the heights of the four points in $\mathcal{P}_{all}$ relative to $\mathbf{F}$ are known as $\mathcal{H}_{all} = \{h_{neck},h_{hip},h_{knee},h_{ankle}\}$. In camera frame $\{\mathbf{C}\}$, the direction of the human's central axis is represented by a unit vector $\mathbf{V_{1}}$ from the optical center $\mathbf{C}$, as shown in Fig. \ref{observation_model}a. Due to the camera's ego-motion when the robot moves on uneven terrain, $\mathbf{V_{1}}$ is not always parallel to the y-axis of camera frame $\{\mathbf{C}\}$ but has 2-DoF rotation from it, which is defined as the rotation from $\mathbf{V_{0}} = (0,-1,0)^{T}$ to $\mathbf{V_{1}}$, marked as $\mathcal{R}_{V_{0}\longrightarrow V_{1}}$.

Furthermore, we assume that a human is always upright while walking. As shown in Fig. \ref{observation_model}b where a quadruped robot is following a person with camera ego-motion, we consider the robot and the target person both to move on a virtual plane that is perpendicular to $\mathbf{V_{1}}$ with footprint $\mathbf{F}$ and robot frame \(\{\mathbf{R}\}\) on it. The rotation from $\{\mathbf{C}\}$ to $\{\mathbf{R}\}$ is exactly $\mathcal{R}_{V_{0}\longrightarrow V_{1}}$, which can be represented by either rotation matrix $\mathbf{R}$ or Euler angles $\{\theta, \phi\}$ in z-x-y order: 
\begin{equation}
    \mathbf{R} = \mathbf{R_z(\phi)R_x(\theta)},
\end{equation}
where $\mathbf{R_z(\phi)}$ and $\mathbf{R_x(\theta)}$ are elementary rotation matrices of the roll and the pitch angles, respectively.

When the person and the robot start moving, the camera rotation $\{\theta, \phi\}$ and the camera height $h_{C}$ (the length of segment CR in Fig. \ref{observation_model}b) 
 varies due to camera ego-motion, while the person's location $(X_{F}, Z_{F})$ on the virtual plane changes and needs to be estimated for velocity control mentioned in Sec. \ref{framework}. The parameters of the system comprise a state vector $\mathbf{s}$:

\begin{equation}
    \mathbf{s}= \{X_{F}, Z_{F}, h_{C}, \mathbf{\theta}, \mathbf{\phi}\}
\end{equation}

Now we derive constraints between the state vector $\mathbf{s}$ and the measurements $\mathcal{N}_{all}$ through the camera projection model.

In robot frame $\{\mathbf{R}\}$, vector $\overrightarrow{\mathbf{CP_{i}}}$ from camera optical center $\mathbf{C}$ to a point $\mathbf{P_{i}} \in \mathcal{P}_{all}$ is: 

\begin{equation}
\label{5}
    \overrightarrow{\mathbf{CP_{i}}} = (X_{F} , h_{C} - h_i, Z_{F})^T, h_i \in \mathcal{H}_{all}
\end{equation}

Therefore, we have:
\begin{equation}
    \mathbf{P_{i}^C} = \mathbf{R^{-1}} \cdot \overrightarrow{\mathbf{CP_{i}}},
\end{equation}   
where $\mathbf{P_{i}^C}$ is the coordinates of $\mathbf{P_{i}}$ in camera frame $\{\mathbf{C}\}$.

\subsection{Optimization Details} \label{optimization}
As people walk, each of the four points in $\mathcal{P}_{all}$ deviates from the assumed rigid model with varying magnitudes. To account for this, we empirically assign different weights to each point. Specifically, lower weights are allocated to highly mobile points ($P_{knee}$ and $P_{ankle}$) compared to more stable upper-body points ($P_{neck}$ and $P_{hip}$).

Here we calculate the weighted reprojection error $f$:
\begin{equation}
    f(X_F, Z_F, h_C, \theta, \phi) = \sum_{i=1}^{n} w_i \left\| \mathbf{n_{i}} - \pi(\mathbf{P_{i}^C}) \right\|^2
\label{repro_1}
\end{equation}
where $\pi$ is the camera projection function, $\pi: \mathbb{R}^3 \rightarrow \mathbb{R}^2$, which projects $\mathbf{P_{i}^C}$ onto the normalized image plane. $w_i$ is the corresponding weight for each point.

By minimizing $f$, we can optimize variables in  $\mathbf{s}$. Specifically, we divide $\mathbf{s}$ into two parts: the translation part $\mathbf{t} = \{X_F, Z_F, h_C\}$ and the rotation part $\mathbf{r} = \{\theta,\phi\}$. We alternately update one part while fixing the other as Pavlakos\cite{6-dof} did to ease the parameter coupling:

\[
\begin{aligned}
&\text{Repeat until convergence:} \quad
\left\{
\begin{aligned}
\mathbf{t^*} &\leftarrow \arg\min_{\mathbf{t}} f(\mathbf{t}, \mathbf{r}) \\
\mathbf{r^*} &\leftarrow \arg\min_{\mathbf{r}} f(\mathbf{t}, \mathbf{r})
\end{aligned}
\right.
\end{aligned}
\]

This is a nonlinear least-squares problem with a small number of variables and a dense Jacobian. We can constrain the variables in the state vector $\mathbf{s}$ with bounds based on the real motion limits of the camera and the person. We employ Dogbox\cite{dogbox}, an extension of Powell’s dogleg method\cite{dogleg} to solve this bounded nonlinear least-squares problem. To enhance the robustness of the reprojection error against noise, we employ the Cauchy cost function\cite{ba}. The initial values of the state vector $\mathbf{s}$ are set around a level viewing state at each frame.

\subsection{Implementation in RPF Framework} \label{framework}

\begin{figure*}
        \centering
        \includegraphics[width=1.0\linewidth]{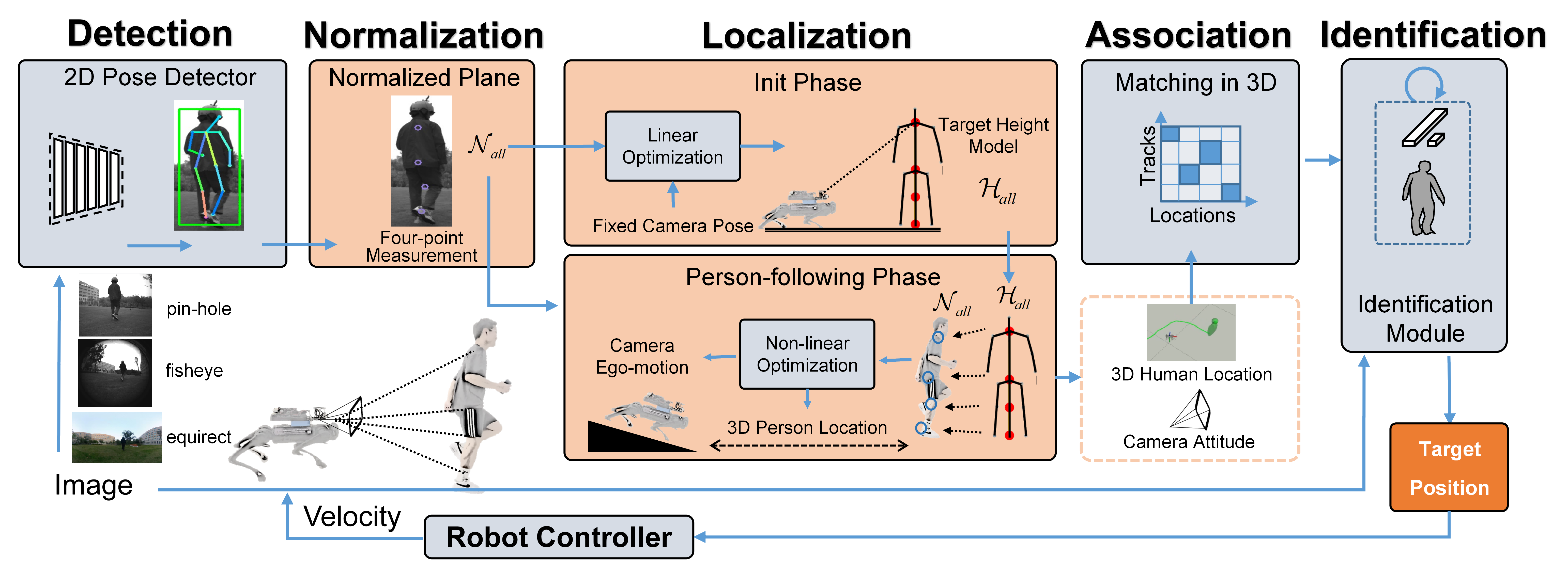}
        \caption{Our proposed framework for monocular Robot Person Following (RPF). The modules highlighted in orange represent our key contributions: (1) a normalization step for camera-agnostic processing, and (2) a subsequent optimization-based person localization method.}
        \label{method-framework}
        \vspace*{-0.25in}
\end{figure*}

Our Robot Person Following (RPF) framework is illustrated in Fig. \ref{method-framework}. Building upon our previous work\cite{ye2023robot}, it follows a standard pipeline \cite{islam2019person,koide2020monocular,Bacchin-Italian,ye2023robot}, which takes a monocular image stream as input and produces velocity control commands as output.

The pipeline begins with detecting human bounding boxes and 2D joints from the input image. We back-project the measurements into the normalized image plane to make our method generalizable to different cameras (e.g., pinhole, fisheye, and equirectangular). Subsequently, the normalized measurements are fed into the person localization module, which estimates the person's footprint location on the virtual plane defined in Fig. \ref{method-framework} by optimization.

The localization module contains two phases. In the initialization phase, the robot estimates the target person's joint heights $\mathcal{H}_{all}$ in a static state (e.g., a quadruped lying on the ground) with known camera attitude $\{\theta, \phi\}$ and height $h_{C}$. Therefore the reprojection error in Eq. \ref{repro_1} is formulated as:
\begin{equation}
    g(X_F, Z_F, \mathcal{H}_{all}) = \sum_{i=1}^{n} w_i \left\| \mathbf{n_{i}} - \pi(\mathbf{P_{i}^C}) \right\|^2
\end{equation}
Minimizing this objective function $g$ allows us to jointly optimize for the person's location $(X_{F}, Z_{F})$ and their joint heights $\mathcal{H}_{all}$. This is a linear least squares problem that can be efficiently solved using SVD\cite{svd}:
\begin{equation}
    {X_F^*, Z_F^*, \mathcal{H}_{all}^*} = \arg\min_{X_F, Z_F, \mathcal{H}_{all}} g(X_F, Z_F, \mathcal{H}_{all})
\end{equation}
In the subsequent person-following phase, the robot utilizes the calibrated joint heights $\mathcal{H}_{all}$ to simultaneously estimate the person's real-time location and the camera's attitude (see Sec. \ref{optimization}).

Following the per-frame localization, a data association step links new detections to existing 3D tracks, each of which is then updated using a Kalman filter. To ensure the robot follows the correct individual, a person re-identification (re-ID) module \cite{re-id} is employed to identify the designated target among all active tracks. Finally, the control module navigates the robot to the person-following position based on the target person's location $(X_{F}, Z_{F})$.

\section{EXPERIMENTS}
\subsection{Baselines}

To evaluate the localization accuracy of the proposed method, we conduct a comparison with geo-model-based and deep-learning-based methods introduced in Sec.~\ref{geometric} and Sec. ~\ref{regress}, respectively. The baselines are named as follows:
\textit{Geo-model-based:} 
\begin{itemize}
        \item \textbf{Koide's Method}\cite{koide2020monocular} locates the person by the neck observation under the assumption of a fixed camera.
        \item \textbf{Ye's Method}\cite{ye2023robot} extends Koide's method to use observations of four points separately to handle occlusions, still restricted to a fixed camera.
\end{itemize}
\textit{Deep-learning-based:} 
\begin{itemize}
        \item \textbf{MonoLoco++}\cite{monoloco++} estimates the 3D coordinates of the bounding box's center of the person in camera frame by a neural network trained on the KITTI dataset\cite{kitti}.  
        \item \textbf{Depth Anything}\cite{depth_anything} outputs relative depth map of the scene. We use the ground truth depth of the person in the first scene as the metric, and compute the average depth of all the joints as the person's distance from the camera.
       
        \item \textbf{Multi-HMR}\cite{multi-hmr2024} focuses on human mesh recovery but also estimates the absolute distance between the pelvis (the center of human hips) and the camera through a neural network.
\end{itemize}

\subsection{Datasets \label{dataset-text}}
A limited number of datasets target the evaluation of person localization from a robot's view with camera ego-motion. To prove the generalizability of our method, we conduct experiments on two public datasets:

\begin{itemize}
    \item \textbf{FieldSAFE}\cite{fieldsafe}, which is the closest to our scenario to the best of our knowledge. This dataset consists of approximately two hours of multi-sensor data collected from a tractor-mounted system in a grass-mowing scenario and provides precise localization of the observed moving pedestrians.

    \item \textbf{KITTI}\cite{kitti} where most deep-learning-based person localization models are trained on. However, it should be noted that the KITTI dataset, which covers autonomous driving scenarios, does not include data distributions with severe camera ego-motion from a robot's view.

\end{itemize}

To evaluate the proposed method and to support further research on this topic, we constructed and released a dataset named RPF-Quadruped.

\begin{figure}[H]
    \centering
    \begin{subfigure}{.5\linewidth}
        \centering
        \includegraphics[width=0.75\linewidth]{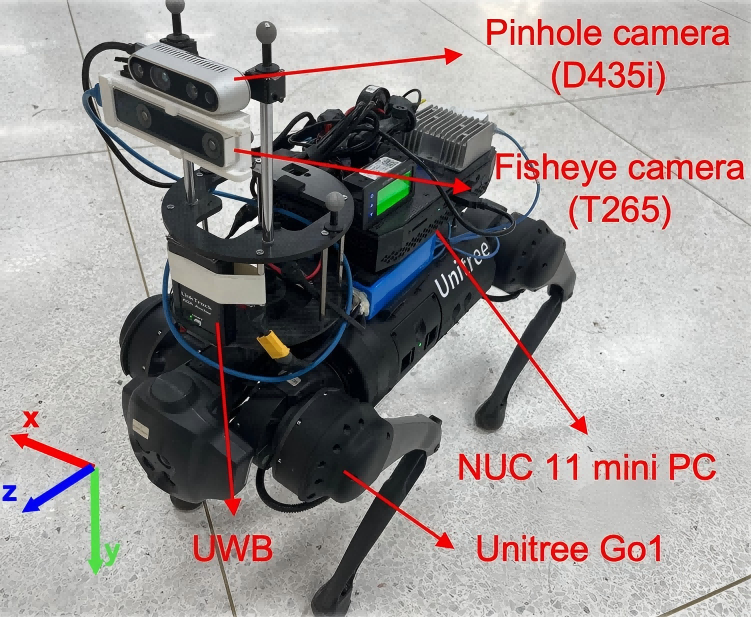}
        \caption{Platform}
        \label{platform}
    \end{subfigure}%
    \begin{subfigure}{.5\linewidth}
        \centering
        \includegraphics[width=0.75\linewidth]{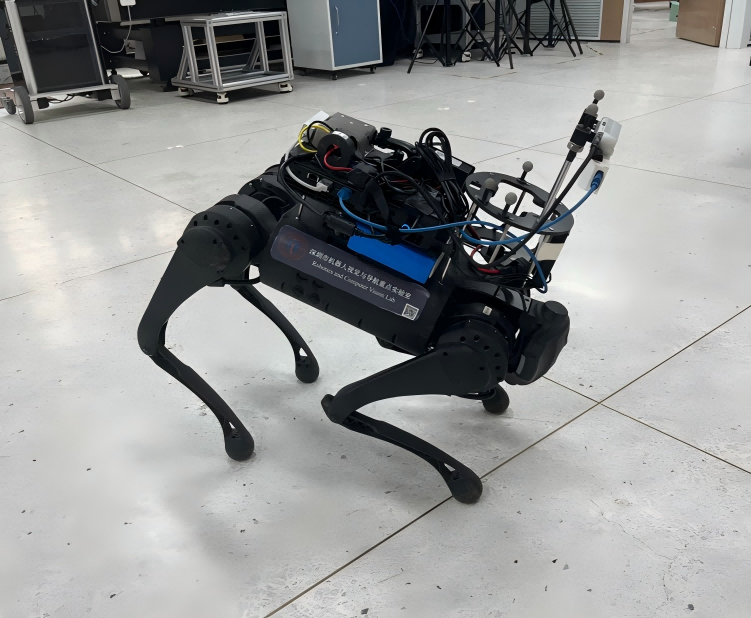}
        \caption{Turning Head}
        \label{Turning Head}
    \end{subfigure}
    
    \vspace{0.5ex} 
    
    \begin{subfigure}{.5\linewidth}
        \centering
        \includegraphics[width=0.8\linewidth]{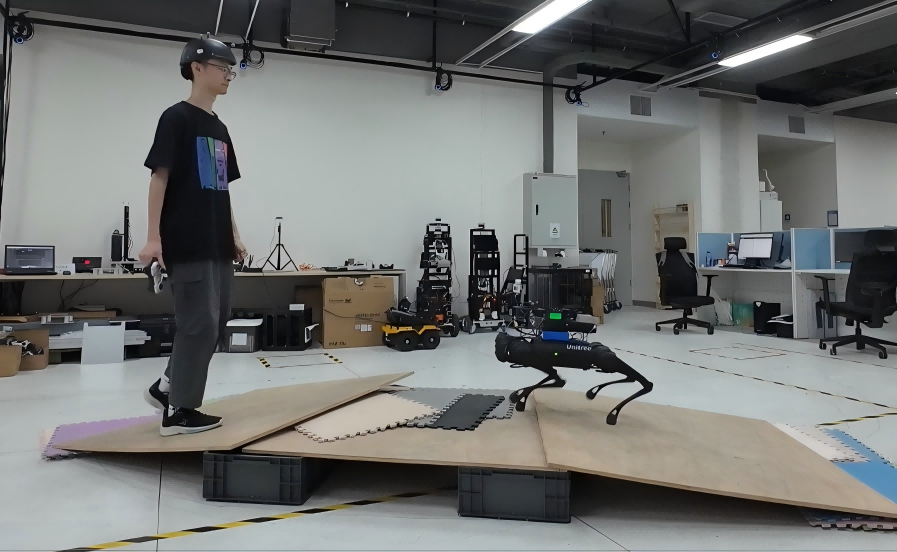}
        \caption{Indoor Slope}
        \label{Indoor Slope}
    \end{subfigure}%
    \begin{subfigure}{.5\linewidth}
        \centering
        \includegraphics[width=0.8\linewidth]{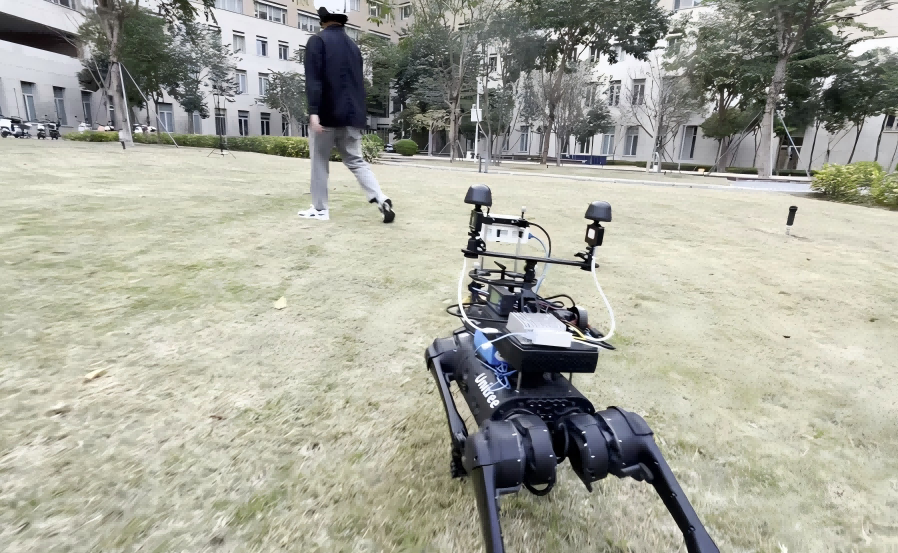}
        \caption{Rugged Lawn}
        \label{Rugged Lawn}
    \end{subfigure}
    \caption{(a) Our quadruped robot platform. (b-d) Scenarios from our RPF-Quadruped dataset.}
    
    \label{fig:2x2grid}
    \vspace*{-0.15in}
\end{figure}

RPF-Quadruped dataset is recorded from the quadruped robot Unitree Go1\cite{unitree}, illustrated in Fig. \ref{platform} (refer to Sec. \ref{platform-text}). It consists of three scenarios: \textit{Turning Head}, \textit{Indoor Slope}, and \textit{Rugged Lawn}. Fig. \ref{Turning Head}, \ref{Indoor Slope}, \ref{Rugged Lawn} are snapshots of them, respectively. In the latter two scenarios, we run our RPF system in real time while collecting data from pin-hole, fisheye, and panoramic cameras. In the \textit{Rugged Lawn} scenario, the Go1 robot followed the target person for six minutes over a distance of 354 meters without any pauses. We utilized UWB to measure the person's distance from the camera as the ground truth. In the first two scenarios, we used a motion capture system to record the relative poses between the person and the robot as ground truth.

RPF-Quadruped dataset includes data collected at close range, with large-angle ego-motion and continuous observations of a real walking person from a low viewing angle. Table \ref{dataset_comparison} illustrates the range of the data distribution compared to other datasets.

\begin{table}[h!]
\centering
\scalebox{0.67}{
\begin{tabular}{@{}lcccccc@{}}
\toprule
\textbf{Dataset}       & \textbf{KITTI\cite{kitti}}       & \textbf{FieldSAFE\cite{fieldsafe}}          & \textbf{RPF-Quadruped}\\ \midrule
\textbf{Distance from Camera (m)} & 18.44 $\pm$ 11.20          & 7.50 $\pm$ 1.50         & 3.50 $\pm$ 3.00 \\ 
\textbf{Camera Height (m)} & 2.31 $\pm$ 0.29        & 4.50 $\pm$ 0.09           &  0.50 $\pm$ 0.15 \\ 
\textbf{Camera Pitch (deg)} &  /       &      16.01 $\pm$ 5.27      & 0.5 $\pm$ 15.46\\ 
\textbf{Camera Roll (deg)} &  /       &       0.32 $\pm$ 4.18     & 0.8 $\pm$ 10.30\\
\end{tabular}
}
\caption{Statistical comparison of key parameters across different datasets. The table reports the \textbf{mean} and \textbf{standard deviation} for person-to-camera distance, camera height, and camera ego-motion (pitch and roll angles). Note that the KITTI dataset\cite{kitti} lacks pitch and roll data.}
\label{dataset_comparison}
\end{table}

\begin{figure}[H]
    \centering
    \begin{subfigure}{.5\linewidth}
        \centering
        \includegraphics[width=0.95\linewidth]{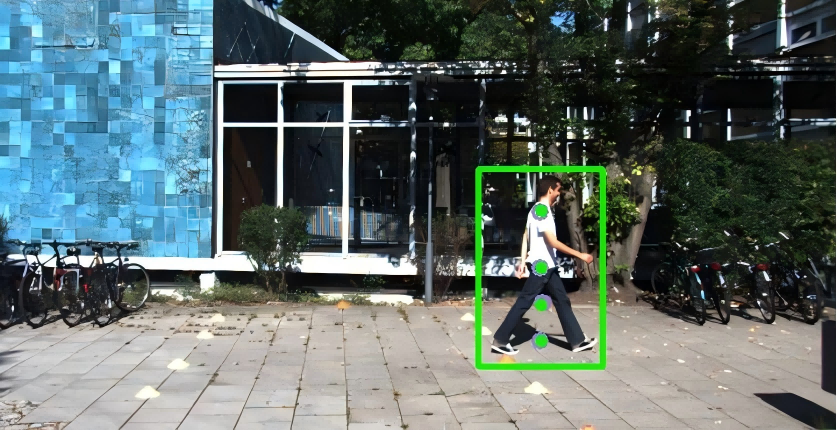}
        \caption{KITTI}
        \label{fig:kitti}
    \end{subfigure}%
    \hfill
    \begin{subfigure}{.5\linewidth}
        \centering
        \includegraphics[width=0.95\linewidth]{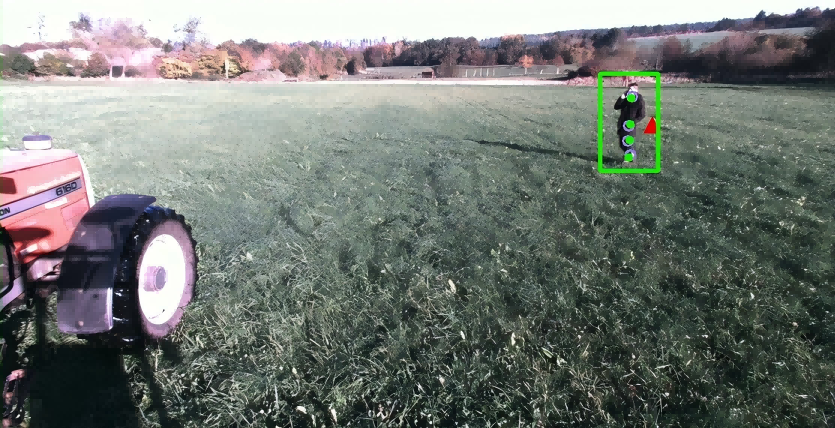}
        \caption{FieldSAFE}
        \label{fig:fieldsafe}
    \end{subfigure}
    
    \begin{subfigure}{.5\linewidth}
        \centering
        \includegraphics[height = 0.56\linewidth]{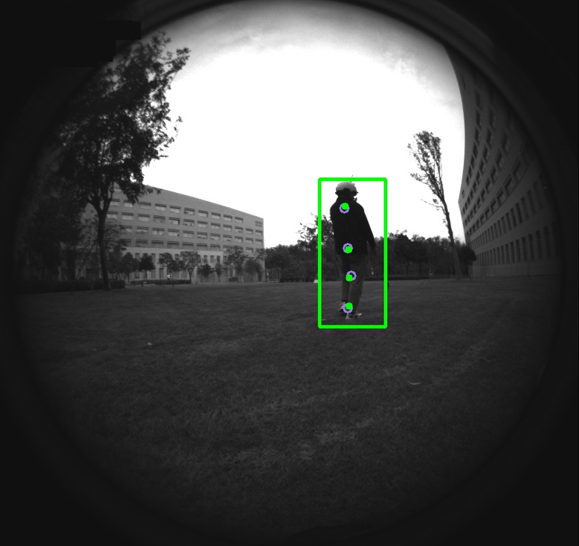}
        \caption{Rugged Lawn (Fisheye)}
        \label{fig:fisheye}
    \end{subfigure}%
    \begin{subfigure}{.5\linewidth}
        \centering
        \includegraphics[height = 0.56\linewidth]{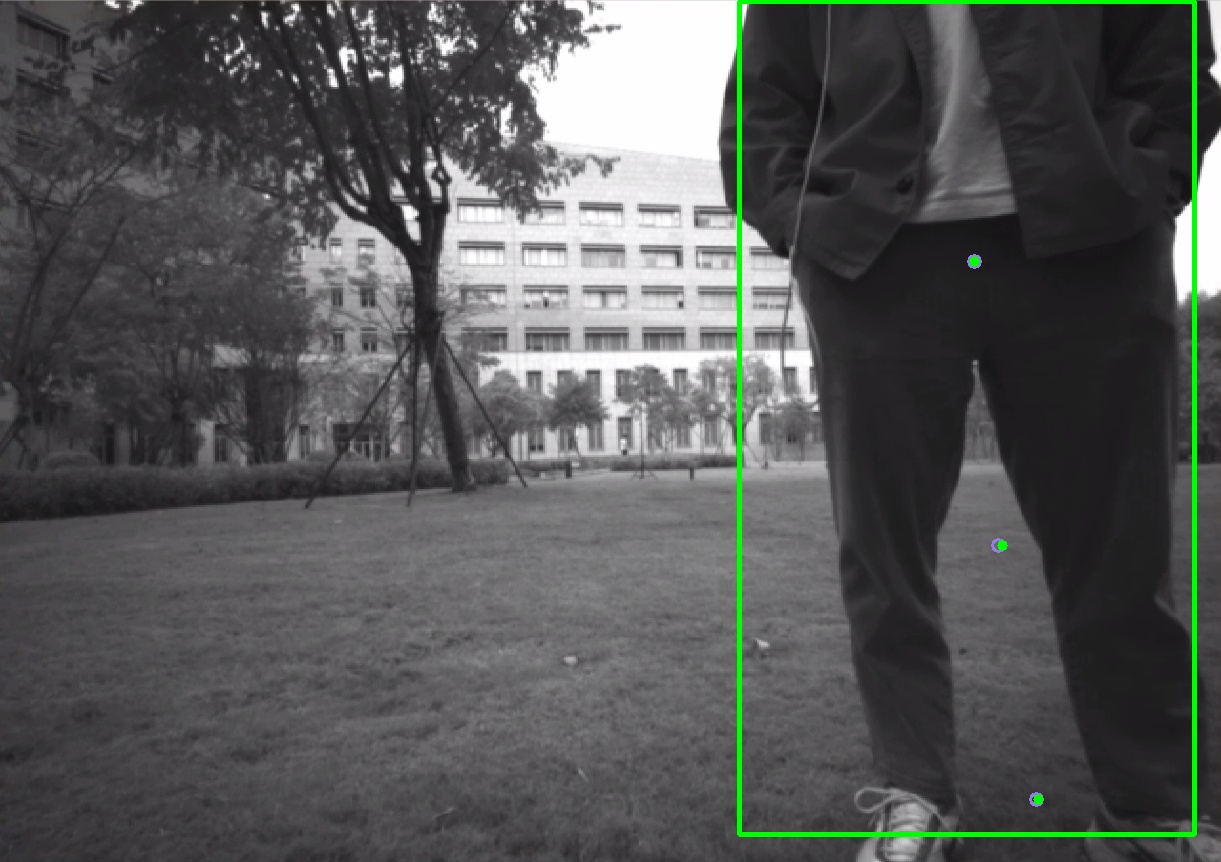}
        \caption{Rugged Lawn (Pin-hole)}
        \label{fig:pin-hole}
    \end{subfigure}
    \caption{Screenshots of our method running on different datasets: (a) KITTI\cite{kitti}, (b) FieldSAFE\cite{fieldsafe}, (c-d) fisheye and pin-hole images in \textit{Rugged Lawn}. The neck point in (d) is not observable due to the person's proximity to the camera.}
    \label{fig:Screenshots}
    \vspace*{-0.15in}
\end{figure}

\subsection{Platform and Implementation Details\label{platform-text}}

We use a quadruped robot Unitree Go1\cite{unitree} as our experimental platform (shown in Fig. \ref{platform}), which is equipped with: an onboard mini PC Intel NUC 11 containing CPU i7-1165g7 and GPU RTX 2060-mobile (6G), a mounted UWB distance sensor with 0.1m accuracy (Nooploop LinkTrack AOA) for ground truth distance measurement, and two forward-facing cameras (Intel RealSense D435i and T265), capturing 640*480 pin-hole and 848*800 fisheye grayscale images with global shutter at 30 Hz, respectively. 

Due to the limited space on Go1, another Ricoh Theta S panoramic camera can be mounted, replacing the two Intel cameras, as shown in Fig. \ref{introduction}. It publishes 1280*720 equirectangular images at 15 Hz.

It is worth mentioning that, compared to the quadruped platforms used in existing RPF methods\cite{zhang2021efficient, Multi-modal, 3D-pose-korean}, our Go1 robot is smaller in size and has a higher step frequency. As a result, it performs severe ego-motion in rough terrain. 

We evaluate all the methods except for Multi-HMR\cite{multi-hmr2024} on the onboard mini PC of the Go1 robot. We adopt the largest model for Depth Anything\cite{depth_anything} and Multi-HMR\cite{multi-hmr2024}. For methods that use 2D human joint detection as input, the measurements are uniformly obtained through the method specified in Sec. \ref{mod and obs}. We leverage NVIDIA's TensorRT technology to accelerate the inference of YOLOX\cite{yolox} and AlphaPose\cite{fang2022alphapose} for real-time performance. As for Multi-HMR, due to the massive model parameter size, the mini PC is unable to run it; instead, we conduct the testing offline on a separate PC equipped with RTX 3070 (8G) and an Intel i7-12700F.
\subsection{Evaluation and Results}

We evaluate the accuracy and runtime of person localization methods, which are essential for achieving reliable person localization in real-time robotic systems.

For localization accuracy, we uniformly adopt the pelvis (the center of human hips) in the camera frame to represent the 3D location of a human. Additionally, we calculate the distance of the human relative to the optical center, which is equal to the L2 norm of the location. We employ the \textbf{Average Location Error (ALE)} and \textbf{Average Distance Error (ADE)} to evaluate the accuracy of localization as \cite{monoloco++,ye2023robot} did. Moreover, for FieldSAFE\cite{fieldsafe} and \textit{Rugged Lawn} scenario in our dataset with continuous camera ego-motion, we introduce the \textbf{Variance of Location Error (VLE)} and \textbf{Variance of Distance Error (VDE)} as additional metrics to assess the stability of localization. Since all the baselines only support the pin-hole camera model, we use the pin-hole images in our dataset for evaluation across baselines. Fig. \ref{fig:kitti}, \ref{fig:fieldsafe} show screenshots of our method running on public datasets. Fig. \ref{fig:fisheye} and Fig. \ref{introduction} show our method running on equirectangular and fisheye images, respectively. Fig. \ref{fig:pin-hole} shows our method still works under partial observation (discussed in Sec. \ref{discuss}).

The results on localization accuracy are shown in Table \ref{PL-table}. Our method achieves the lowest average error and variance across our dataset and FieldSAFE dataset. However, in KITTI dataset, MonoLoco++ demonstrates higher accuracy since it is trained on this dataset.

\begin{figure}[H]
        \centering
        \includegraphics[width=0.9\linewidth, height=0.5\linewidth]{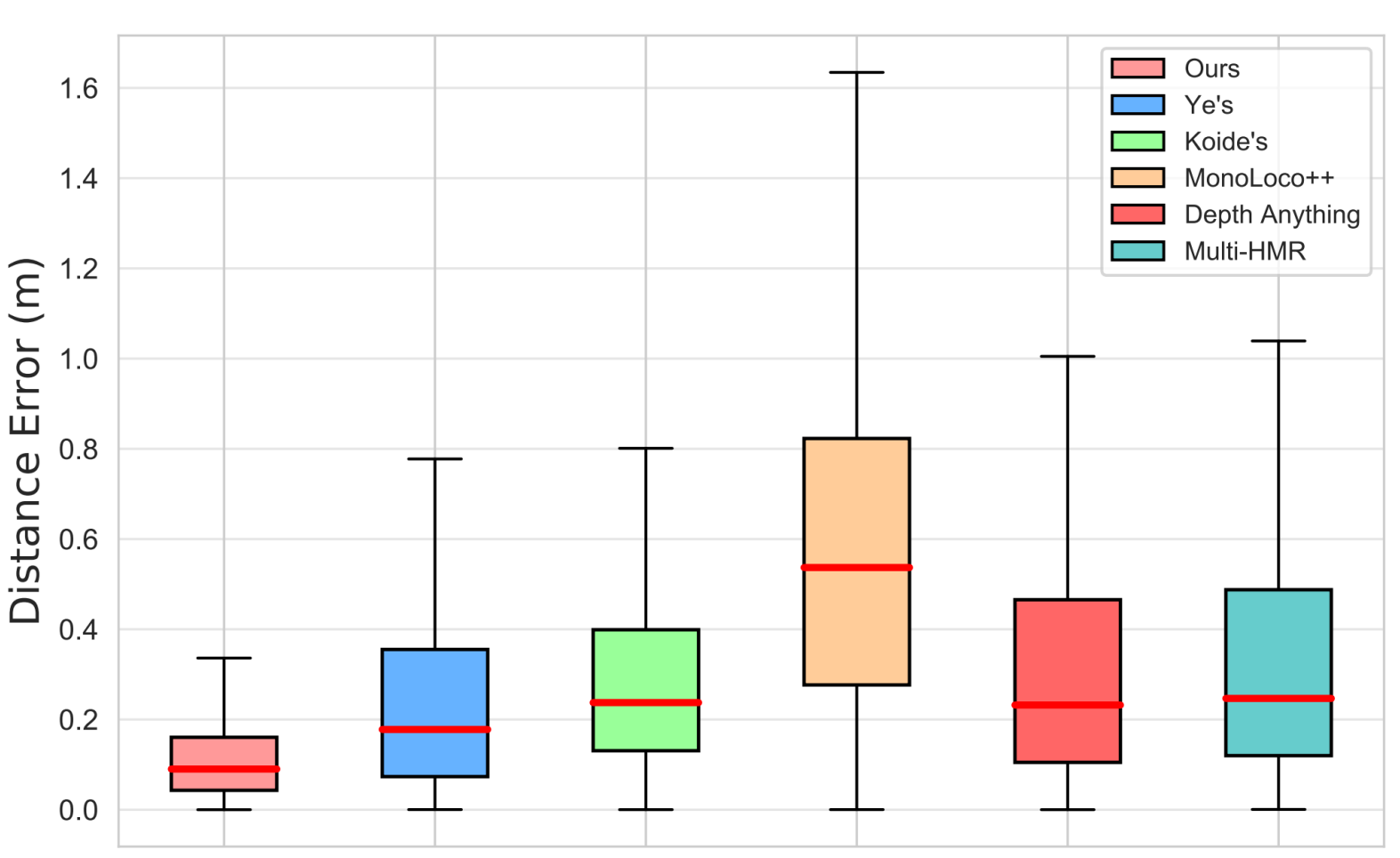}
        \caption{A box plot illustrating the distance error of our method compared to baselines in \textit{Rugged Lawn} scenario. Our method achieves more accurate and stable distance estimation compared to the baseline methods.}
        \label{box}
        \vspace*{-0.15in}
\end{figure}

\begin{figure}[H]
        \centering
        \vspace*{-0.15in}
        \includegraphics[width=0.92\linewidth, height=0.75\linewidth]{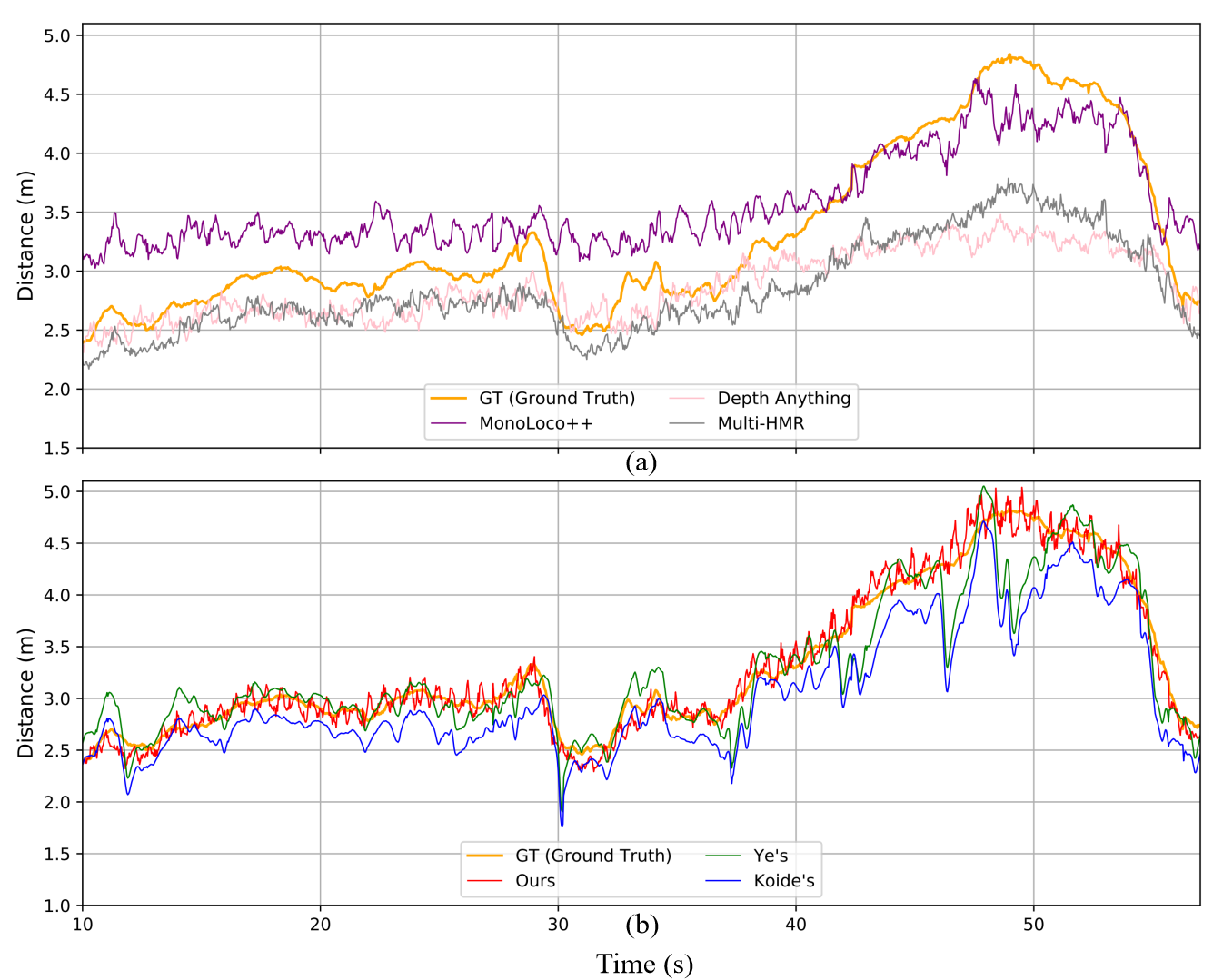}
        \caption{Comparison of estimated distance over time on a sequence from the \textit{Rugged Lawn} dataset (10s--57s). The plot shows the output of (a) deep-learning-based and (b) geo-model-based baselines. Our method's estimate demonstrates a higher fidelity to the ground truth than baselines. }
        
        \label{result_draft}
        \vspace*{-0.15in}
\end{figure}

Additionally, a box plot illustrating the distance error of our method compared to baselines in the \textit{Rugged Lawn} scenario is presented in Fig. \ref{box}. Our method demonstrates more accurate and stable distance estimation throughout the six-minute RPF process, even under continuous and severe camera ego-motion. Specifically, we extract a clip of the \textit{Rugged Lawn} scene (from 10s to 57s) and plot the person distance estimation curves for both deep-learning-based (Fig. \ref{result_draft}(a)) and geo-model-based methods (Fig. \ref{result_draft}(b)). 

\begin{table}[hb]
\vspace*{-0.1in}
        \centering
    \caption{\upshape{Comparison of localization accuracy. We evaluate our method against several baselines and present an ablation study. Metrics include: Average Location/Distance Error (\textbf{ALE}/\textbf{ADE}) in meters (m), and their corresponding variances (\textbf{VLE}/\textbf{VDE}) in m\textsuperscript{2}. Variances are reported only for datasets with sequential frames (\textit{Rugged Lawn} and FieldSAFE \cite{fieldsafe}). For the \textit{Rugged Lawn} dataset, distance-based metrics (ADE/VDE) are used to align with its UWB-based ground truth of distance.}}
	\scalebox{0.52}{
		\begin{tabular}{l|c|c|c|c|c}
			\toprule
                        \multirow{2}*{\diagbox[]{\textbf{Methods}}{\textbf{Scenarios}}} 

                        &\multicolumn{1}{c|}{\textbf{Turning Head}} &\multicolumn{1}{c|}{\textbf{Indoor Slope}} &\multicolumn{1}{c|}{\textbf{Rugged Lawn}} &\multicolumn{1}{c|}{\textbf{FieldSAFE\cite{fieldsafe}}}&\multicolumn{1}{c}{\textbf{KITTI\cite{kitti}}}\\

			&\textbf{\textit{ALE}} $\downarrow$
            &\textbf{\textit{ALE}} $\downarrow$
            &\textbf{\textit{ADE}} / \textbf{\textit{VDE}} $\downarrow$
            &\textbf{\textit{ALE}} / \textbf{\textit{VLE}} $\downarrow$
            &\textbf{\textit{ALE}} $\downarrow$\\

                        \midrule

                        Koide's Method\cite{koide2020monocular} & 0.396  & 0.289  & 0.3 / 0.3 & 1.924 / 5.012 & 1.451  \\
                                                
                        Ye's Method\cite{ye2023robot}    & 0.294  &  0.261 & 0.3 / 0.3 & 1.856 / 3.952 & 1.420 \\   

                        MonoLoco++\cite{monoloco++}     & 0.820  & 0.510  & 0.6 / 0.2 & 4.152 / 4.705 & \textbf{0.940} \\

                        Depth Anything\cite{depth_anything} & 0.571  & 0.523  & 0.5 / 0.6 & 1.528 / 1.022 & 2.963 \\

                        Multi-HMR\cite{multi-hmr2024}       &  0.493 & 0.254  & 0.4 / 0.3 &  3.066 / 0.424 & 1.520\\
                        
                        \textbf{Ours} 
                            & \textbf{0.178}  & \textbf{0.101}  & \textbf{0.1} / \textbf{0.0} &  \textbf{1.287} / \textbf{0.356} & 1.220\\
                        \cmidrule(lr){1-6}
                        Ours w/o neck
                            & 0.238  & 0.196  & 0.2 / 0.1 &  1.324 / 0.865 & 1.320\\
                        Ours w/o ankle
                            & 0.204  & 0.141  & 0.1 / 0.0 &  1.308 / 0.401 & 1.275\\
                        Ours on fisheye images
                            & 0.182  & 0.119  & 0.1 / 0.0 &  / & /\\ 
			\bottomrule
	\end{tabular}}
	\label{PL-table}
\end{table}

As shown in Fig. \ref{result_draft}(a), due to weak generalization, the deep-learning-based methods struggle to accurately reflect the trend of person distance changes during continuous RPF. This is particularly evident with Monoloco++\cite{monoloco++}, which was trained on the KITTI dataset\cite{kitti} that lacks data distribution for a person below three meters from the camera, making it difficult to predict accurate distances of a close person in \textit{Rugged Lawn}. From Fig. \ref{result_draft}(b), the geo-model-based methods generally capture the trend of distance changes. However, when the camera undergoes large-angle ego-motion, methods except ours have significant errors, which could pose safety concerns for an RPF system that relies on localization for robot speed control. This further highlights the robustness and necessity of our method.

\begin{table}[h]
\centering
\vspace*{-0.05in}
\caption{Comparison of per-frame average runtime. The preprocessing time accounts for 2D human joint detection. This step is not required by Depth Anything\cite{depth_anything} and Multi-HMR\cite{multi-hmr2024}, which derive location directly from depth maps and 3D meshes, respectively. *Runtime for Multi-HMR was measured on a different PC (see Sec. \ref{platform-text}).}
\scalebox{0.7}{
\begin{tabular}{@{}lccc@{}}
\toprule
Method & Preprocessing (s) & Estimation (s) & Total (s) \\ \midrule
Koide's Method\cite{koide2020monocular} & 0.02 & \textbf{0.0006} & \textbf{0.0206}  \\
Ye's Method\cite{ye2023robot} & 0.02 & 0.0008 & 0.0208  \\
MonoLoco++\cite{monoloco++} & 0.02 & 0.09 & 0.11  \\ 
Depth Anything\cite{depth_anything} & / & 0.23 & 0.23  \\
Multi-HMR\cite{multi-hmr2024}$^*$ & / & 1.24 & 1.24  \\
Ours & 0.02 & 0.005 & 0.025  \\ \bottomrule
\end{tabular}
}
\label{timecomparison}
\vspace*{-0.15in}
\end{table}

For runtime performance, the results are shown in Table \ref{timecomparison}. Compared to methods based on simple geometric models for planar scenarios, our method is slightly slower but still ensures strong real-time performance. In contrast to deep-learning-based methods, our method significantly outperforms in terms of time efficiency.

\subsection{Discussion \label{discuss}}

\begin{itemize}
    \item \textbf{Partial observation}. Our method still works even when only three points  in $\mathcal{P}_{all}$ are observed, demonstrated in Fig. \ref{fig:pin-hole}. As shown in Table \ref{PL-table}, when the neck's or ankle's observation is removed across datasets, our method maintains high localization accuracy. 
    \item \textbf{Performance against posture change}. As shown in Fig. \ref{result_draft}(b), the distance estimated by our method consistently oscillates around the ground truth at a certain frequency. This oscillation arises because the human body undergoes periodic deformation during walking. Such fluctuations in distance can be easily smoothed out by a filter. However, when the human breaks the upright assumption, such as sitting down, our method fails as most geo-model-based methods do. This issue can be engineerly resolved by incorporating a human posture recognition module that filters out non-upright postures.
    
\end{itemize}

\section{CONCLUSIONS}

In this paper, we propose an optimization-based method to locate a target person under camera ego-motion. By representing an upright human with a four-point model, our method jointly estimates camera attitude and person location. A Robot Person Following (RPF) system is proposed based on our method that achieves real-time (40 FPS on an onboard mini PC) monocular person localization and tracking from an agile quadruped's view in rough terrains. We create and make public a dataset recorded from the quadruped to supplement the research of monocular person localization under camera ego-motion, which is critical for human-robot interaction such as RPF\cite{islam2019person} and egocentric surveillance systems\cite{anomaly}. Experiments on both public datasets and our dataset demonstrate the effectiveness of our method compared to deep-learning-based and geo-model-based baselines. Future work will involve extending our method by employing expressive human models to handle a wider range of postures. The localization accuracy could also be improved by explicitly estimating the ground plane on which the person stands. Extensive evaluations can be conducted on large-scale egocentric datasets with dense crowds, such as TPT-bench\cite{tpt-bench}.




\bibliographystyle{bibliography/IEEEtran}
\bibliography{bibliography/refs}

\end{document}